\documentclass[lettersize,journal]{IEEEtran}
\usepackage[T1]{fontenc}
\usepackage{amsmath,amssymb,amsfonts}
\usepackage{algorithmic}
\usepackage{algorithm}
\usepackage{array}
\usepackage[caption=false,font=normalsize,labelfont=sf,textfont=sf]{subfig}
\usepackage{textcomp}
\usepackage{stfloats}
\usepackage{url}
\usepackage{verbatim}
\usepackage{graphicx}
\usepackage[table]{xcolor}
\usepackage{bbm}
\usepackage{booktabs}
\usepackage{multirow}
\usepackage{arydshln}
\usepackage{cite}
\usepackage{arydshln}
\newcommand{\posdelta}[2]{\cellcolor{blue!#1}{$\uparrow\,#2$}}
\newcommand{\negdelta}[2]{\cellcolor{red!#1}{$\downarrow\,#2$}}
\newcommand{\basedelta}{\cellcolor{gray!10}{--}}
\usepackage{pifont}
\usepackage{makecell}

\begin{document}

\title{Principle-Guided Supervision for Interpretable Uncertainty in Medical Image Segmentation}

\author{An~Sui$^{\dagger}$,
        Yuzhu~Li$^{\dagger}$,
        Gunter~Schumann,
        Fuping~Wu$^*$,
        and~Xiahai~Zhuang$^*$%
\thanks{A. Sui and X. Zhuang are with the School of Data Science, Fudan University, Shanghai, China.}
\thanks{Y. Li and G. Schumann are with the Institute of Science and Technology for Brain-Inspired Intelligence, Fudan University, Shanghai, China.}
\thanks{F. Wu is with the National Heart and Lung Institute, Imperial College London, London, United Kingdom.}
\thanks{$^\dagger$ A. Sui and Y. Li contributed equally to this work.}
\thanks{$^*$ X. Zhuang and F. Wu are the corresponding authors.}
}

\markboth{IEEE TRANSACTIONS ON PATTERN ANALYSIS AND MACHINE INTELLIGENCE, Under Review}
{Sui \MakeLowercase{\textit{et al.}}: Principle-Guided Supervision for Interpretable Uncertainty in Medical Image Segmentation}


\maketitle

\begin{abstract}
Uncertainty quantification complements model predictions by characterizing their reliability, which is essential for high-stakes decision making such as medical image segmentation. However, most existing methods reduce uncertainty to a scalar confidence estimate, leaving its spatial distribution semantically underconstrained.
In this work, we focus on \emph{uncertainty interpretability}, namely, whether estimated uncertainty behaves in a human-understandable manner with respect to sources of ambiguity. We identify three perception-aligned principles requiring the spatial distribution of uncertainty to reflect: (1) image contrast between structures, (2) severity of image corruption, and (3) geometric complexity in anatomical structures. Accordingly, we develop a principle-guided uncertainty supervision framework (PriUS) based on evidential learning, in which the corresponding supervision objectives are explicitly enforced during training. We further introduce quantitative metrics to measure the consistency between predicted uncertainty and image attributes that induce ambiguity. Experiments on ACDC, ISIC, and WHS datasets showed that, compared with state-of-the-art methods, PriUS produced more consistent uncertainty estimates while maintaining competitive segmentation performance.
\end{abstract}

\begin{IEEEkeywords}
Uncertainty supervision, uncertainty interpretability, medical image segmentation.
\end{IEEEkeywords}

\section{Introduction}\label{sec:introduction}

\IEEEPARstart{D}{eep} learning models have achieved remarkable performance across a wide range of visual recognition tasks \cite{khadem2026role, erdur2024deep}. In medical image analysis, however, accuracy alone is insufficient for safe deployment in clinical workflows \cite{csahin2025unlocking, singh2025interpretability}. Practical use requires not only a prediction, but also an assessment of how reliable that prediction is and where potential failures may arise. This demand has made uncertainty estimation an important component of trustworthy medical image segmentation \cite{lambert2024trustworthy}.

A broad range of uncertainty estimation methods has been developed for dense prediction, including Bayesian inference \cite{gal2016dropout}, deep ensembles \cite{lakshminarayanan2017simple}, test-time augmentation (TTA) \cite{wang2019aleatoric}, and evidential deep learning (EDL) \cite{sensoy2018evidential}. These methods provide useful confidence signals and have demonstrated value in reliability assessment. Nevertheless, in many existing approaches, uncertainty is primarily treated as a scalar estimate and evaluated solely by calibration \cite{ECE} or by its correlation with prediction errors \cite{UEO}. As a result, the spatial distribution of uncertainty often remains semantically underconstrained. Although a model may assign high uncertainty to particular image regions, it is often unclear whether such uncertainty meaningfully reflects the underlying sources of ambiguity.

This issue is particularly important in high-stakes scenarios, such as medical image segmentation, where ambiguity is rarely arbitrary \cite{baumgartner2019ambiguity}. In practice, it often arises from identifiable image characteristics, including weak boundary contrast \cite{thompson1980contrast}, severe image corruption \cite{kamann2021corruption}, and complex anatomical geometry \cite{kronman2016geometric}. 
These factors directly affect the difficulty of delineating a target structure. Conventional uncertainty estimation methods, however, do not explicitly constrain uncertainty estimates to follow the corresponding expected responses. Consequently, uncertainty maps may indicate low confidence while failing to reveal whether their spatial patterns are aligned with the image properties that make predictions difficult.

These limitations motivate the study of \emph{uncertainty interpretability}:
whether the spatial variation of uncertainty is consistent with \emph{why} a prediction is difficult, rather than merely indicating \emph{how much} it can be trusted. This perspective shifts the focus from scalar confidence estimation to the semantic spatial behavior of uncertainty.
To this end, we build on our preliminary work \cite{li2025uncertainty}, and identify three perception-aligned principles characterizing how predictive uncertainty varies with (1) boundary contrast, (2) image corruption, and (3) anatomical geometry.
We then develop a \textbf{Pri}nciple-guided \textbf{U}ncertainty \textbf{S}upervision framework (PriUS), which operationalizes each principle using a measurable image-derived proxy and a corresponding supervision objective. To evaluate uncertainty interpretability beyond segmentation accuracy, we further introduce quantitative metrics that measure uncertainty-ambiguity consistency. 


The main contributions of this work are threefold:

\begin{itemize}
 \item We formulate \emph{uncertainty interpretability} for medical image segmentation through three perception-aligned principles. These principles specify how predictive uncertainty should vary with boundary contrast, image corruption, and anatomical geometry.

We propose PriUS, a unified principle-guided uncertainty supervision framework instantiated with evidential learning. PriUS translates the three principles into supervision objectives using measurable proxies, including image gradient, noise level, and boundary distance.

 \item We introduce quantitative metrics for uncertainty interpretability. Experiments on multiple benchmarks demonstrated that PriUS produced uncertainty estimates that better reflected ambiguity-inducing image properties, while maintaining competitive segmentation performance.
\end{itemize}

The rest of this paper is organized as follows. 
Section~\ref{sec:related_work} reviews related literature on uncertainty estimation, uncertainty calibration, and evidential learning. 
Section~\ref{sec:methods} presents the proposed PriUS, from the formulation of principles to the uncertainty supervision objective and its evidential learning instantiation.
Section~\ref{sec_exp} evaluates PriUS on three medical image segmentation benchmarks and analyzes its uncertainty interpretability and segmentation performance. 
Finally, Section~\ref{sec:conclusion} concludes this paper.

\section{Related Work}\label{sec:related_work}

Our work draws on three related lines of research: \textit{uncertainty estimation} (Section~\ref{subsec_ue}), which provides the predictive uncertainty that our framework seeks to make more interpretable; \textit{uncertainty calibration} (Section~\ref{subsec_uc}), which pursues improved reliability but does not explicitly enforce perceptually meaningful spatial behavior; and \textit{evidential deep learning} (Section~\ref{subsec_edl}), which serves as the uncertainty prediction backbone in our framework owing to its single-pass efficiency and explicit evidence parameterization.

\subsection{Uncertainty Estimation}
\label{subsec_ue}

Uncertainty in deep learning is commonly categorized into data (aleatoric) and model (epistemic) uncertainty \cite{HORA, kendall2017uncertainties}. Data uncertainty arises from inherent ambiguity in the observations, such as imaging corruption, low contrast, or indistinct anatomical boundaries, and cannot be eliminated even with additional training data. By contrast, model uncertainty reflects uncertainty in the learned parameters due to limited or insufficient training samples and can be reduced as more data become available. In addition, recent studies have discussed distributional uncertainty, which refers to uncertainty induced by distributional mismatch between training and testing data \cite{ulmer2021prior}.

Bayesian deep learning methods approximate predictive uncertainty by sampling model outputs under stochastic perturbations \cite{gal2016dropout}. Such methods are mainly used to capture model uncertainty by reflecting the sensitivity of predictions to uncertainty in the learned parameters. Ensemble-based approaches further improve uncertainty estimation by aggregating predictions from multiple independently trained models, yielding more robust uncertainty measures at the cost of increased computational and memory requirements \cite{lakshminarayanan2017simple,rahaman2021uncertainty,liu2019accurate}. Similar to Bayesian approximations, ensemble methods are commonly associated with model uncertainty, as disagreement among models reflects incomplete knowledge of the true underlying mapping. TTA \cite{wang2019aleatoric} offers a lightweight alternative that infers uncertainty from prediction variability under different input perturbations. Under augmentation designs that simulate observation-level degradations, such variability can reflect sensitivity to local image corruption. Accordingly, TTA-derived uncertainty is often associated with data uncertainty and also serves as a practical proxy for predictive instability. Beyond Bayesian and sampling-based formulations, deterministic approaches estimate uncertainty through simple proxies such as entropy \cite{namdari2019review}, variance \cite{willems2012model}, and softmax-derived confidence scores \cite{ovadia2019can}, or through evidential deep learning \cite{sensoy2018evidential}. While these approaches offer practical and computationally efficient indicators of predictive uncertainty, they do not explicitly distinguish between different uncertainty sources.

\subsection{Uncertainty Calibration}
\label{subsec_uc}

Beyond uncertainty estimation, a growing body of work investigates how to calibrate uncertainty so that it better align with empirical correctness.
Mehrtash \textit{et al.} \cite{mehrtash2020confidence} studied predictive uncertainty together with confidence calibration and showed that ensemble-based uncertainty estimation improved calibration quality, out-of-distribution detection, and the overall reliability of dense predictions. Zhang \textit{et al.} \cite{zhang2020mix} proposed Mix-n-Match to improve the expressiveness and data efficiency of post-hoc calibration while preserving the predictive accuracy of the base model. Zou \textit{et al.} \cite{zou2023towards} proposed DEviS, which explicitly modeled calibrated evidential uncertainty for reliable medical image segmentation, improving robustness, calibration, and inference efficiency within a single-pass framework. Sun \textit{et al.} \cite{rednet} further improved the reliability of multimodal medical image segmentation by discounting unreliable modality-specific evidence during evidential fusion, thereby refining the trustworthiness of the resulting uncertainty-aware predictions.

Compared with uncertainty calibration, explicit supervision of uncertainty has received less attention. For example, Abutalip \textit{et al.} \cite{EDUE} proposed EDUE, which leveraged disagreement among expert annotators to guide one-pass uncertainty estimation in medical image segmentation. By aligning predicted uncertainty with genuine ambiguity reflected by multi-rater annotations, EDUE improved both calibration and the correspondence between uncertainty and expert disagreement. However, the supervisory signal in such methods is primarily driven by annotation disagreement, and explicit supervision of uncertainty according to human-interpretable perceptual or structural principles remains largely underexplored.

\subsection{Evidential Deep Learning}
\label{subsec_edl}

EDL has recently emerged as an efficient alternative for predictive uncertainty modeling, particularly in dense prediction tasks such as medical image segmentation. Grounded in Dempster-Shafer theory \cite{shafer1992dempster} and subjective logic \cite{josang2016subjective}, EDL represents model outputs as evidence parameters of a Dirichlet distribution, enabling simultaneous estimation of class probabilities and predictive uncertainty within a single deterministic forward pass \cite{ovadia2019can,zou2023towards,deng2023uncertainty}. Unlike Bayesian sampling or ensemble-based methods, EDL directly parameterizes a higher-order distribution over categorical probabilities, with uncertainty quantified by the amount of supporting evidence.

By modeling evidence strength explicitly, EDL provides a unified representation of predictive uncertainty that reflects both intrinsic data ambiguity and insufficient evidence in uncertain regions. However, most existing EDL-based approaches focus primarily on the quality of uncertainty quantification, calibration, or robustness under distributional variations \cite{gao2025edlsurvey}. Comparatively little attention has been paid to explicitly constraining how uncertainty should behave spatially with respect to perceptually meaningful ambiguity cues in medical images.

\begin{figure*}[!t]
 \centering
 \includegraphics[width=0.95\linewidth]{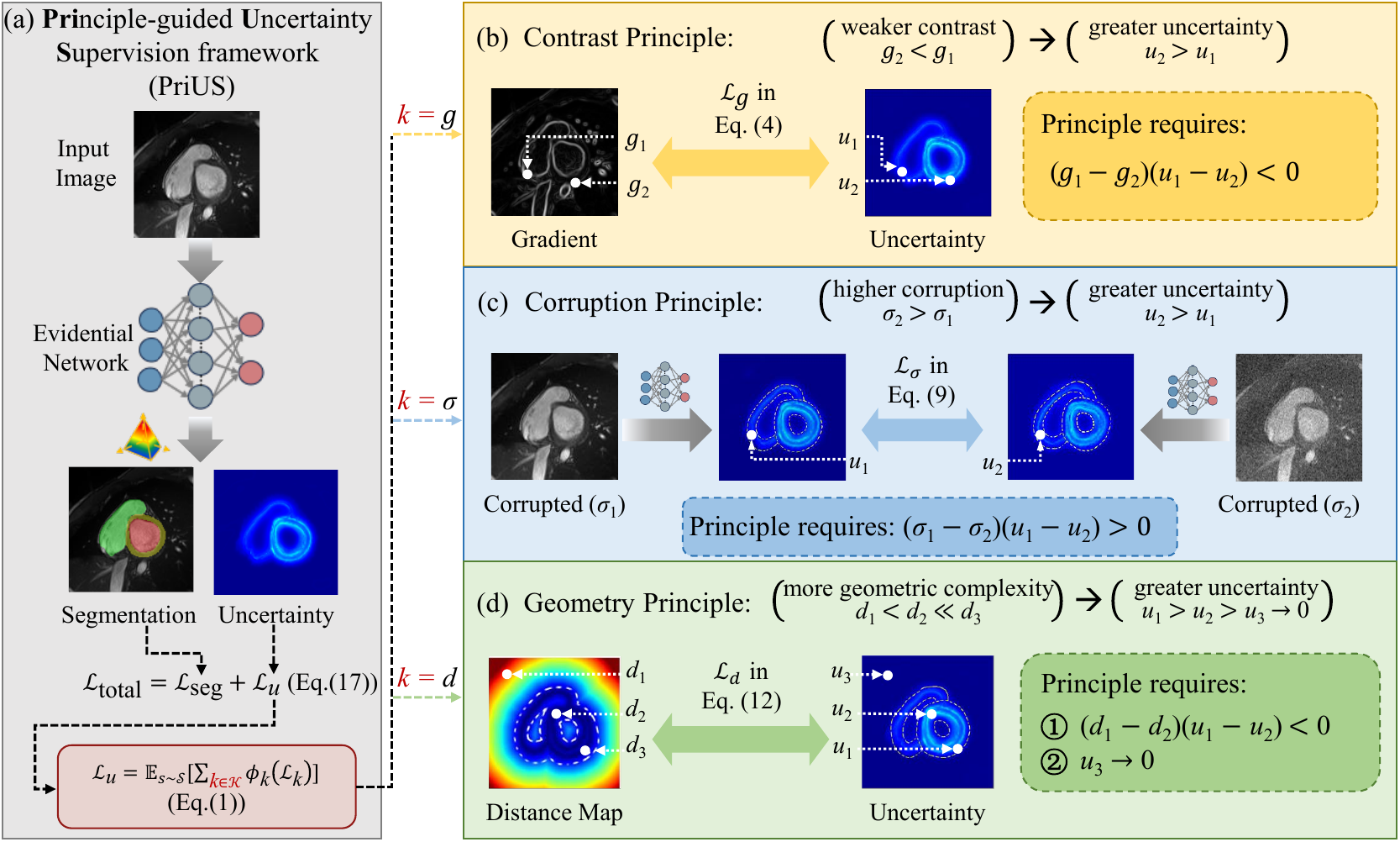}
 \caption{Illustration of the proposed Principle-guided Uncertainty Supervision framework (PriUS) and its three principle-specific supervision designs. \textbf{(a)} PriUS uses an evidential network to produce segmentation predictions and uncertainty estimates, and optimizes a joint objective that combines segmentation loss with principle-guided uncertainty supervision. \textbf{(b)} Contrast Principle ($k=g$): lower boundary contrast should correspond to higher uncertainty. \textbf{(c)} Corruption Principle ($k=\sigma$): stronger corruption should correspond to higher uncertainty. \textbf{(d)} Geometry Principle ($k=d$): geometrically complex regions should exhibit higher uncertainty, and uncertainty should decrease progressively toward spatially coherent interiors far from boundaries.}
 \label{fig:overview}
\end{figure*}

\section{Methods}\label{sec:methods}

To improve the interpretability of predictive uncertainty, we propose PriUS, a principle-guided uncertainty supervision framework. Section~\ref{subsec_principles} identifies three perception-aligned principles characterizing how predictive uncertainty should vary with image properties that contribute to segmentation ambiguity. Section~\ref{subsec:sup_loss} formulates a unified supervision objective and instantiates it through principle-specific losses, each combining a base supervision term with a gating function derived from the corresponding ambiguity proxy. Section~\ref{subsec:evidential} describes the evidence-based uncertainty prediction module and presents the overall training framework of PriUS.

\subsection{Perception-Aligned Principles}
\label{subsec_principles}

To address the lack of explicit structural supervision on uncertainty, we formulate uncertainty interpretability with respect to perceptually meaningful sources of uncertainty. 
Structural interpretation is shaped by the image contrast between adjacent regions, the reliability of the observed signal, and the spatial transition from boundary regions to coherent anatomical interiors. 
Strong and mutually consistent evidence tends to support confident prediction, whereas weak, degraded, or conflicting evidence increases uncertainty. 
Based on this observation, we establish three perception-aligned principles, namely the Contrast Principle, the Corruption Principle, and the Geometry Principle.


\par\vspace{7pt}
\textit{Contrast Principle:}
\textit{Predictive uncertainty should be lower in regions with stronger contrast between adjacent structures.}
\par\vspace{7pt}

Structural inference relies on the discriminability between neighboring regions. When the contrast between adjacent structures is strong, visual evidence supports confident prediction. In contrast, weak or indistinct transitions reduce discriminability and increase ambiguity.

\par\vspace{7pt}
\textit{Corruption Principle:}
\textit{Predictive uncertainty should be higher in regions where image quality is more severely degraded.
}
\par\vspace{7pt}

Predictive reliability depends on the fidelity of the observed signal. Noise, artifacts, and other perturbations degrade the quality of the observation, weakening the evidence relevant to structural prediction and thereby increasing predictive uncertainty.

\par\vspace{7pt}
\textit{Geometry Principle:}
\textit{Predictive uncertainty should be higher in geometrically complex regions and lower in spatially coherent ones.}
\par\vspace{7pt}

Predictive uncertainty is also shaped by anatomical geometry, particularly the spatial distance to semantic boundaries. 
Pixels near anatomical boundaries tend to admit multiple plausible delineations, whereas pixels farther inside spatially coherent regions support more confident label assignments.

These principles characterize three complementary sources of perceptual ambiguity, including image contrast, observational corruption, and structural geometry, which together provide a unified perspective for modeling uncertainty in structured visual prediction tasks.

\subsection{Principle-Guided Uncertainty Supervision}\label{subsec:sup_loss}

To operationalize the proposed principles, we formulate a unified uncertainty supervision objective. It regularizes predicted uncertainty according to image-derived proxies, which serve as measurable surrogates for the underlying factors that contribute to segmentation ambiguity.
Each principle-specific term consists of two components: a base supervision loss that encodes the desired uncertainty relation induced by the corresponding proxy, and a gating function that modulates this supervision according to local structural context.
The overall objective is therefore written as
\begin{equation}
\mathcal{L}_u
=
\mathbb{E}_{s\sim\mathcal{S}}
\left[
\sum_{k\in\mathcal{K}}
\phi_k\left(s;\mathcal{L}_k(u_s; z_s^{(k)})\right)
\right],
\label{eq:unified_lu}
\end{equation}
where $\mathcal{S}$ denotes a sampling distribution over spatial samples $s$. 
Here, $s$ can be a pixel index $i$ or a pixel index pair $(i,j)$ depending on the instantiated principle. 
The notation $u_s$ compactly denotes the uncertainty value(s) involved in $s$, e.g., $u_i$ when $s=i$, and $(u_i,u_j)$ when $s=(i,j)$.
The set $\mathcal{K}=\{g,\sigma,d\}$ contains the uncertainty principles considered in the framework, corresponding to contrast, corruption, and geometry, respectively. 
For each principle $k$, $\mathcal{L}_k(\cdot)$ denotes the base supervision loss, $z_s^{(k)}$ denotes the image-derived proxy associated with $s$, and $\phi_k(\cdot)$ denotes the principle-specific gated supervision term. 
In this way, uncertainty constraints are enforced only when the corresponding proxy provides informative local supervision.

We compute the soft weights in these gated supervision terms using a slope-scaled sigmoid function,
\begin{equation}
    \rho(x)\triangleq \frac{1}{1+\exp(-\gamma x)},
\label{eq:sigmoid}
\end{equation}
where \(x\) is instantiated by distance-related expressions in the following principle-specific gates, and \(\gamma\) controls the transition sharpness.


Under this shared formulation, each principle is instantiated by pairing the corresponding image-derived proxy with its base supervision loss and gating mechanism. In this work, we consider three such instantiations, i.e., image gradient for the Contrast Principle, noise level for the Corruption Principle, and boundary distance for the Geometry Principle.

\paragraph{Contrast Supervision}
Local image gradients provide a direct, model-agnostic surrogate for structural contrast. Strong image gradients typically indicate clear transitions between adjacent regions, whereas weak image gradients correspond to indistinct or smoothly varying boundaries. Based on this observation, we operationalize the Contrast Principle by linking uncertainty supervision to image-gradient-derived contrast strength. Since this principle concerns how uncertainty varies with boundary discriminability, supervision is applied primarily in boundary-relevant regions rather than homogeneous interiors.

Specifically, for each boundary-adjacent pixel \(i\), we denote its spatial coordinate and unit normal direction by \(\mathbf m_i\) and \(\mathbf n_i\). 
We sample pixel indices along the normal direction as
\[
\mathcal{N}_i
=
\{\, \Pi(\mathbf m_i+t\mathbf n_i) \mid t\in\mathcal{T} \,\},
\]
where \(\mathcal{T}=\{-T,\ldots,T\}\) denotes a set of discrete offsets, \(T\) is a fixed sampling radius, and \(\Pi(\cdot)\) maps a spatial coordinate to its nearest valid pixel index. 
Given the pixel-wise image-gradient magnitude \(g_p\) and predicted uncertainty \(u_p\) for $p\in\mathcal{N}_i$, we define the normal-aggregated quantities for pixel \(i\) as
\begin{equation}
\tilde g_i=\max_{p\in\mathcal{N}_i} g_p,
\qquad
\bar u_i=\frac{1}{|\mathcal{N}_i|}\sum_{p\in\mathcal{N}_i}u_p,
\label{eq:normal-aggregation}
\end{equation}
which capture the strongest local image-gradient response and the averaged uncertainty along $\mathbf n_i$, respectively.

Then, given a pixel pair $(i,j)$ belonging to the same semantic class, the Contrast Principle implies that the location with stronger local contrast should exhibit lower uncertainty. The corresponding base supervision loss is defined as
\begin{equation}
\mathcal{L}_{g}((u_i,u_j); z_{ij}^{(g)})
=
\max\bigl(
0,\,
(\bar u_i-\bar u_j)(\tilde g_i-\tilde g_j)
\bigr)
,
\label{eq:gu-base}
\end{equation}
where $z_{ij}^{(g)}=(\tilde g_i,\tilde g_j)$ denotes the contrast proxy pair for pixels $i$ and $j$.
This loss penalizes violations of the desired inverse ordering between contrast and uncertainty, encouraging stronger gradient-supported transitions to receive lower uncertainty than weaker ones.

To restrict this supervision to boundary-proximal regions, we define the corresponding gating function as
\begin{equation}
\phi_g\left((i,j);\mathcal{L}_{g}\right)
=
\omega_g(i,j)\cdot \mathcal{L}_{g},
\label{eq:ggate}
\end{equation}
with
\begin{equation}
\omega_g(i,j)=\lambda_g\cdot\rho\bigl(d_g-\max(d_i,d_j)\bigr),
\label{eq:wg}
\end{equation}
where \(d_i\) and \(d_j\) denote the distances of pixels \(i\) and \(j\) to their nearest semantic boundaries, \(d_g\) controls the spatial extent of the supervised region, and \(\lambda_g\) is a balancing coefficient.
This gating function acts as a soft spatial mask, emphasizing regions near structural transitions while attenuating contributions from interior areas.

\paragraph{Corruption Supervision}
For the Corruption Principle, we introduce corruption through additive Gaussian noise in a controlled and model-agnostic manner. 
Given an input image \(\mathbf{x}\), we construct a family of perturbed observations
\begin{equation}
\mathbf{x}^{[n]} = \mathbf{x} + \boldsymbol{\eta}^{[n]},
\qquad
\boldsymbol{\eta}^{[n]} \sim \mathcal{N}\left(\mathbf{0},(\sigma^{[n]})^2\mathbf{I}\right),
\label{eq:corruption}
\end{equation}
where \(\sigma^{[n]}\) parameterizes the noise level. Larger values of \(\sigma^{[n]}\) correspond to stronger degradation of image fidelity, yielding an ordered spectrum from clean images to increasingly corrupted versions.

The Corruption Principle requires uncertainty to increase monotonically along this spectrum.
For a local region $\mathcal{P}_i$ centered at pixel $i$, we define the aggregated uncertainty under noise level $n$ as
\begin{equation}
\bar u_i^{[n]}
=
\frac{1}{|\mathcal{P}_i|}
\sum_{p\in\mathcal{P}_i}u_p^{[n]},
\label{eq:patch-uncertainty}
\end{equation}
where $n\in\{0,1,2\}$ indexes increasing noise levels, and $u_p^{[n]}$ denotes the predicted uncertainty at pixel $p$ under noise level $n$.

Then we define the base supervision loss as
\begin{equation}
\begin{aligned}
&\mathcal{L}_{\sigma}(u_i; z_i^{(\sigma)})= \\
&
\sum_{n=1}^{2}
\max\bigl(
0,\,
-(\sigma^{[n]}-\sigma^{[n-1]})(\bar u_i^{[n]}-\bar u_i^{[n-1]})
\bigr),
\label{eq:nu-base}
\end{aligned}
\end{equation}
where \(z_i^{(\sigma)}=\{\sigma^{[n]}\}_{n=0}^{2}\) denotes the corruption proxy associated with pixel \(i\).
This loss penalizes violations of monotonic uncertainty growth as corruption increases, encouraging predictive uncertainty to reflect degradation in observational evidence.

To focus corruption-guided supervision on regions where reliability degradation directly affects structural delineation,  we define the corresponding gating function as
\begin{equation}
\phi_{\sigma}\left(i;\mathcal{L}_{\sigma}\right)
=
\omega_{\sigma}(i)\cdot \mathcal{L}_{\sigma},
\label{eq:gsigma}
\end{equation}
with
\begin{equation}
\omega_{\sigma}(i)=\lambda_{\sigma}\cdot\rho(d_n-d_i),
\label{eq:wsigma}
\end{equation}
where \(d_i\) is the distance from pixel \(i\) to the nearest semantic boundary, \(d_n\) controls the spatial extent of supervision, and \(\lambda_{\sigma}\) is a balancing coefficient. 
This gating function emphasizes near-boundary, anatomically meaningful regions while suppressing homogeneous interior areas.

\paragraph{Geometry Supervision}
To instantiate the Geometry Principle, we derive a boundary-distance map in which each pixel value represents the Euclidean distance to the nearest semantic boundary. 
Specifically, object boundaries are first extracted from the segmentation mask, after which a Euclidean distance transform is applied to obtain the distance map.
The resulting distance map provides a continuous surrogate for structural stability: pixels near structural transitions are generally more ambiguous, whereas pixels deeper inside anatomical regions tend to be spatially coherent. 

Accordingly, the geometry-induced prior involves two complementary behaviors: uncertainty should decrease as boundary distance increases, and it should further diminish in sufficiently homogeneous interior regions. 
To encode these behaviors explicitly, we first describe geometry supervision for a sampled pixel pair \((i,j)\) in the following idealized two-regime form:
\begin{equation}
\begin{aligned}
&\mathcal{L}_{d}((u_i,u_j); z_{ij}^{(d)})
 = \\
&\begin{cases}
\max\bigl(0,\,(d_i-d_j)(u_i-u_j)\bigr), & (i,j)\in\mathcal{S}_{b}, \\
\max\left(0,\,u_i+u_j\right), & (i,j)\in\mathcal{S}_{h},
\end{cases}
\end{aligned}
\label{eq:geometry-piecewise}
\end{equation}
where \(z_{ij}^{(d)}=(d_i,d_j)\) denotes the geometry proxy pair for pixels \(i\) and \(j\).
The set \(\mathcal{S}_{b}\) denotes boundary-adjacent pixel pairs, where at least one pixel lies close to a semantic boundary, and supervision on this set is imposed through a ranking constraint.
In contrast, \(\mathcal{S}_{h}\) denotes homogeneous interior pixel pairs, where both pixels lie sufficiently far from semantic boundaries, and uncertainty is jointly suppressed toward zero.

To distinguish these two regimes in a differentiable manner, we first define
\begin{equation}
t_{ij}
=
\rho(d_i-d_f)\cdot \rho(d_j-d_f),
\label{eq:tij}
\end{equation}
where \(d_f\) is a threshold for identifying pixels sufficiently far from their nearest semantic boundaries. 
Thus, \(t_{ij}\) acts as a soft indicator of whether both sampled pixels belong to the homogeneous interior regime.

To emphasize ordering supervision only when the geometric difference between the two pixels is sufficiently discriminative, we further define
\begin{equation}
\omega_{\epsilon}(i,j)
=
\rho(|d_i-d_j|-d_{\epsilon}),
\label{eq:wepsilon}
\end{equation}
where $d_{\epsilon}$ specifies a margin on boundary-distance difference. This term suppresses weak or non-discriminative distance differences while preserving supervision on geometrically distinguishable pairs.

Based on these quantities, we define the adaptive geometry modulation term
\begin{equation}
\Omega_d(i,j)
=
(1-t_{ij})\cdot \omega_{\epsilon}(i,j)(d_i-d_j)
+
\lambda_f t_{ij},
\label{eq:Omega_d}
\end{equation}
where $\lambda_f$ controls the strength of the interior regularization term. Through the soft regime indicator $t_{ij}$, $\Omega_d(i,j)$ adapts the supervision behavior according to whether the sampled pair lies near the boundary or deep inside a homogeneous region. 

Finally, the corresponding gated supervision term is expressed in a unified form as

\begin{equation}
\phi_d\left((i,j);\mathcal{L}_{d}\right)
=
\max\Bigl(
0,\,
\Omega_d(i,j)\cdot
\bigl(
u_i-
(1-2t_{ij})u_j
\bigr)
\Bigr).
\label{eq:du-base}
\end{equation}

Unlike the previous two principles, the gating behavior here is realized implicitly through the adaptive modulation term $\Omega_d(i,j)$ conditioned on the regime indicator $t_{ij}$, rather than through a separate multiplicative weight.

This formulation yields two complementary regimes. When at least one pixel lies near the boundary, we have $t_{ij}\approx 0$, and Eq.~\eqref{eq:du-base} is approximately given by
\[
\max\bigl(
0,\,
\omega_{\epsilon}(i,j)(d_i-d_j)(u_i-u_j)
\bigr)
,
\]
which enforces an inverse monotonic relationship between uncertainty and boundary distance. In this way, geometry supervision jointly promotes boundary-aware uncertainty decay and interior uncertainty convergence. Conversely, when both pixels are far from the boundary, we have $t_{ij}\approx 1$, and Eq.~\eqref{eq:du-base} is approximately given by
\[
\max\bigl(0,\lambda_f(u_i+u_j)\bigr),
\]
which drives uncertainty in homogeneous interior regions toward zero. 

Together, the three principle-specific supervision terms translate perception-aligned ambiguity principles into measurable image-derived constraints, guiding uncertainty learning toward semantically interpretable spatial behavior.

\subsection{Evidence-Based Prediction and Overall Objective}
\label{subsec:evidential}

The proposed supervision strategy is model-agnostic with respect to the underlying uncertainty estimation mechanism, and can in principle be combined with different predictive uncertainty formulations. In this work, we instantiate it within the evidential deep learning framework, which provides a deterministic evidence-based formulation for jointly modeling class probabilities and predictive uncertainty.

Let $\mathbf{X}=(\mathbf{x}_i)\in\mathbb{R}^{V}$ and $\mathbf{Y}=(\mathbf{y}_i)\in\mathbb{R}^{V\times C}$ denote an image and its label, respectively, where $V$ is the number of pixels and $C$ is the number of classes. The evidential network $f_{\theta}$ estimates an evidence map as $\mathbf{E}=f_{\theta}(\mathbf{X})=(\mathbf{e}_i)\in\mathbb{R}^{V\times C}$. According to subjective logic~\cite{josang2016subjective}, for the $i$-th pixel in $\mathbf{X}$, its categorical probability variable $\mathbf{p}_i=(p_{ic})\in[0,1]^C$ is modeled by a Dirichlet distribution $\mathrm{Dir}(\mathbf{p}_i|\boldsymbol{\alpha}_i)$. The concentration parameter $\boldsymbol{\alpha}_i=[\alpha_{i1},\alpha_{i2},\ldots,\alpha_{iC}]$ is formulated as
\[
\boldsymbol{\alpha}_i = \mathbf{e}_i + W\cdot \mathbf{r},
\]
where $\mathbf{r}$ is a precomputed base-rate prior estimated from the class-wise pixel proportions in the training annotations. Unlike the commonly used uniform base rate, this prior reflects empirical class-frequency statistics and therefore incorporates dataset-level class imbalance into the evidential parameterization. Here, $W$ is a constant weight, set to $C$ in this work. Given the collected evidence, the expected class probability and uncertainty of the $i$-th pixel are computed as
\[
\hat{\mathbf{p}}_i=\frac{\boldsymbol{\alpha}_i}{\sum_{j=1}^{C}\alpha_{ij}},
\qquad
u_i=\frac{C}{\sum_{j=1}^{C}\alpha_{ij}}.
\]

The overall training objective combines the segmentation loss with the principle-guided uncertainty supervision:
\begin{equation}
\mathcal{L}_\mathrm{total} = \mathcal{L}_\mathrm{seg} + \mathcal{L}_{u},
\label{eq:overall}
\end{equation}
where $\mathcal{L}_{u}$ is the uncertainty supervision loss defined in Eq.~\eqref{eq:unified_lu}, and $\mathcal{L}_\mathrm{seg}$ is the segmentation objective

\begin{equation}
\mathcal{L}_{\mathrm{seg}}=\lambda_{\mathrm{CE}}\mathcal{L}_{\mathrm{CE}}+\lambda_{\mathrm{Dice}}\mathcal{L}_{\mathrm{Dice}}+\lambda_{\mathrm{KL}}\mathcal{L}_{\mathrm{KL}},
\label{eq:lseg}
\end{equation}
comprising a cross-entropy loss $\mathcal{L}_{\mathrm{CE}}$, a Dice loss $\mathcal{L}_{\mathrm{Dice}}$, and a Kullback–Leibler divergence term $\mathcal{L}_{\mathrm{KL}}$ that discourages evidence accumulation for incorrect classes.
The resulting framework jointly learns segmentation predictions and uncertainty estimates under evidential modeling, while the principle-guided supervision explicitly regularizes the spatial behavior of uncertainty during training.

\section{Experiments}
\label{sec_exp}
We evaluated PriUS on three medical image segmentation benchmarks: the Automated Cardiac Diagnosis Challenge (ACDC) \cite{bernard2018deep}, the International Skin Imaging Collaboration (ISIC) dataset \cite{codella2019skin,tschandl2018ham10000}, and the Whole Heart Segmentation (WHS) challenge \cite{Zhuang2016MSMMA,Zhuang2019MvMM,GAO2023BayeSeg}. Section~\ref{subsec_exp-settings} describes the datasets and implementation details. Section~\ref{subsec_eval_metrics} introduces the evaluation protocol, comprising the proposed uncertainty interpretability metrics alongside conventional segmentation metrics. Sections~\ref{subsec_results} and~\ref{subsec_segmentation} compare uncertainty interpretability and segmentation accuracy against state-of-the-art methods through quantitative and qualitative studies. Finally, Section~\ref{subsec_parameter} reports a sensitivity analysis of the boundary thresholds, and Section~\ref{subsec_ablation} provides an ablation study examining the contribution of each supervision objective.

\subsection{Experimental Settings}
\label{subsec_exp-settings}

\subsubsection{Datasets}
Details of the three public benchmark datasets are given below.

\begin{itemize}
    \item The ACDC dataset contains cine cardiac MR images with annotations for the left ventricle (LV), right ventricle (RV), and myocardium (Myo). 
    We conducted five-fold cross-validation and reported the mean performance across all folds. 
    Each 2D slice was zero-padded in-plane, and then center-cropped around the foreground region to a fixed size of \(128 \times 128\).

    \item The ISIC dataset consists of dermoscopic RGB images with pixel-wise lesion annotations. 
    All images and masks were resized to a fixed spatial resolution of \(256 \times 256\), and binary lesion masks were derived from the original annotations.

    \item The WHS dataset contains volumetric contrast-enhanced cardiac CT images with expert annotations for seven anatomical structures, including the left ventricle, right ventricle, left atrium, right atrium, myocardium, ascending aorta, and pulmonary artery.
    Due to GPU memory constraints, each volume was first resampled to an isotropic spacing of \(2\times2\times2\ \mathrm{mm}\), padded to a fixed spatial size when necessary, and then cropped into overlapping 3D patches of size \(64 \times 64 \times 64\) for training. During inference, full-volume predictions were obtained by aggregating patch-wise outputs through a Gaussian-weighted sliding-window strategy.
    
\end{itemize}

\subsubsection{Implementation Details}
We adopted U-Net as the segmentation backbone for all methods and optimized the network using Adam with an initial learning rate of \(10^{-3}\). The batch size was set to 24 for ACDC, 32 for ISIC, and 12 for WHS.

For the segmentation loss in Eq.~\eqref{eq:lseg}, we set \(\lambda_{\mathrm{CE}}=1\) and used an annealed Kullback-Leibler regularization weight
\[
\lambda_{\mathrm{KL}}=\min(1,t/20),
\]
where \(t\) denotes the current training epoch. The weight of Dice loss was defined as \(\lambda_{\mathrm{Dice}}=1-\alpha\), with the annealing factor \(\alpha\) given by
\[
\alpha=\alpha_0 \cdot \exp\left(-\frac{\ln \alpha_0}{T}t\right),
\]
where \(\alpha_0=0.01\) and \(T\) is the total number of training epochs.

The supervision coefficients in Eq.~\eqref{eq:wg}, Eq.~\eqref{eq:wsigma}, and Eq.~\eqref{eq:Omega_d} were configured in a dataset-dependent manner. For ACDC, we set
\[
\lambda_g=\lambda_{\sigma}=\alpha,\qquad \lambda_f=100\alpha.
\]
For ISIC, we set
\[
\lambda_g=10\alpha,\qquad \lambda_{\sigma}=\alpha,\qquad \lambda_f=100\alpha.
\]
For WHS, we set
\[
\lambda_g=\alpha,\qquad \lambda_{\sigma}=0.01\alpha,\qquad \lambda_f=10\alpha.
\]

The boundary-range parameter \(d_g\) in Eq.~\eqref{eq:wg}, which controls the spatial extent of contrast-aware supervision, was fixed at 2. For corruption and geometry supervision, the distance thresholds in Eq.~\eqref{eq:wsigma} and Eq.~\eqref{eq:Omega_d} were set symmetrically around a dataset-specific reference threshold \(d_0\), namely \(d_n=d_0-\delta\) and \(d_f=d_0+\delta\), where \(\delta\) is a small positive margin. We set \(d_0=8\) for ACDC, \(d_0=16\) for ISIC, and \(d_0=2\) for WHS. The slope parameter $\gamma$ in Eq.~\eqref{eq:sigmoid} was set to \(100\), yielding sufficiently sharp transitions around the predefined thresholds. All experiments were conducted on a single NVIDIA GeForce RTX 3090 GPU.

\subsection{Evaluation Metrics}
\label{subsec_eval_metrics}
\paragraph{Uncertainty Interpretability Metrics}
We evaluate uncertainty interpretability by measuring how well predicted uncertainty agrees with measurable image-derived proxies of segmentation ambiguity. Intuitively, locations with weaker boundary contrast, stronger corruption, or less stable geometric context should exhibit higher predictive uncertainty. Based on this intuition, we introduce two complementary metrics: (a) the Uncertainty Correlation Coefficient (UCC) and (b) the Uncertainty Ratio (UR).

We define UCC as the Spearman correlation coefficient \cite{wissler1905spearman} between predicted uncertainty and image-derived attributes reflecting local ambiguity. For the Contrast Principle, image gradient magnitude serves as a model-agnostic surrogate for boundary contrast. We therefore compute UCC between the image gradient and the uncertainty over pixels within a boundary region of the \(k\)-th class, defined as \(\text{UCC}_{[g]}=\mathrm{SCorr}(g,u)\) evaluated over \(\mathcal{B}_k(d_0)=\{i \mid y_i=k,\; d_i \le d_0\}\). Specifically, 
\begin{equation}
\begin{aligned}
 &\text{UCC}_{[g]} = \\
 &\frac{\sum_{i \in \mathcal{B}_k(d_0)} \bigl(R(g_i)-\overline{R(g)}\bigr)\bigl(R(u_i)-\overline{R(u)}\bigr)}
 {\sqrt{\sum\limits_{i \in \mathcal{B}_k(d_0)} \bigl(R(g_i)-\overline{R(g)}\bigr)^2
 \sum\limits_{i \in \mathcal{B}_k(d_0)} \bigl(R(u_i)-\overline{R(u)}\bigr)^2}}.
\end{aligned}
\end{equation}
Here, \(R(\cdot)\) denotes the ranking function and \(\overline{(\cdot)}\) denotes the mean value computed over the evaluated pixel set. Similarly, we define \(\mathrm{UCC}_{[\sigma]}=\mathrm{SCorr}(\sigma,u)\) using noise level \(\sigma\) for the Corruption Principle, and \(\mathrm{UCC}_{[d]}=\mathrm{SCorr}(d,u)\) using distance to the nearest semantic boundary \(d\) for the Geometry Principle.

\textit{The UCC value ranges within \([-1, 1]\), with the sign indicating the correlation direction and the magnitude its strength.} Interpretable uncertainty estimates are expected to exhibit a negative correlation with boundary contrast (\(\text{UCC}_{[g]} < 0\)), a positive correlation with corruption severity (\(\text{UCC}_{[\sigma]} > 0\)), and a negative correlation with boundary distance (\(\text{UCC}_{[d]} < 0\)).

UR further quantifies interpretability by measuring the proportion of pixel pairs whose relative uncertainty ordering is consistent with the corresponding image-derived ambiguity. For the Contrast Principle, UR is defined as
\begin{equation}
 \text{UR}_{[g]}=
 \frac{\sum_{i,j \in \mathcal{B}_k(d_0),\, i\neq j}
 \mathbbm{1}_{((g_i-g_j)(u_i-u_j)\leq 0)}}
 {\sum_{i,j \in \mathcal{B}_k(d_0)}\mathbbm{1}_{(i\neq j)}}.
\end{equation}
Similarly, \(\text{UR}_{[\sigma]}\) and \(\text{UR}_{[d]}\) are defined analogously for the Corruption Principle and Geometry Principle over the same pixel set. Higher UR values indicate stronger consistency between uncertainty ordering and ambiguity ordering.

For each metric, computation was first performed at the image level (or volume level for WHS) and then averaged over the evaluation set. For multi-class datasets, the reported values were further averaged across all foreground classes.

\paragraph{Segmentation Accuracy Metrics}
Segmentation performance was evaluated using the Dice similarity coefficient (DSC) and the 95th percentile Hausdorff distance (HD95). For multi-class datasets, the reported values were averaged across all foreground classes.

\subsection{Uncertainty Estimation and Interpretability}
\label{subsec_results}

We compared PriUS with several representative uncertainty estimation approaches, including the EDL-based DEviS \cite{zou2023towards}, the variational inference-based Probabilistic U-Net (PU) \cite{eslami2018probabilistic}, the deep ensemble-based EU \cite{lakshminarayanan2017simple}, the dropout-based UDrop \cite{kendall2017uncertainties}, and TTA \cite{wang2019aleatoric}. For a fair comparison, all baselines were implemented using the same data splits and preprocessing pipeline.

We assessed uncertainty interpretability from both quantitative and qualitative perspectives,
focusing on whether uncertainty estimates consistently align with intrinsic image ambiguity.

\paragraph{Quantitative analysis}
Our analysis proceeds from an overall comparison to a principle-specific examination.
Table~\ref{tab:ucc_ur_comparison} summarizes the UCC and UR results of all methods. 

\begin{table}[t]
 \centering
 \caption{Comparison of uncertainty interpretability metrics UCC and UR. The symbols $(-)$ and $(+)$ indicate that the expected sign of UCC should be negative or positive, respectively. UCC values that violate the expected sign are shaded in gray.}
 \begin{tabular}{ccccccc}
 \toprule
 \multirow{2}{*}{Methods} 
 & \multicolumn{3}{c}{UCC} 
 & \multicolumn{3}{c}{UR} \\
 \cmidrule(lr{0pt}){2-4} \cmidrule(lr{0pt}){5-7}
 & {$g(-)$}$\downarrow$ & {$\sigma(+)$}$\uparrow$ & {$d(-)$}$\downarrow$ 
 & {$g$}$\uparrow$ & {$\sigma$}$\uparrow$ & {$d$}$\uparrow$ \\
 \midrule
 \multicolumn{7}{c}{{ACDC}} \\
 \midrule
 {PriUS} 
 & \textbf{-0.597} & \textbf{0.973} & \textbf{-0.767}
 & \textbf{0.708} & \textbf{0.986} & \textbf{0.739} \\
 {DEviS} 
 & \textcolor{gray}{0.104} & \textcolor{gray}{-0.020} & -0.320
 & 0.463 & 0.490 & 0.516 \\
 {PU} 
 & \textcolor{gray}{0.062} & 0.881 & -0.573
 & 0.478 & 0.937 & 0.721 \\
 {UDrop} 
 & \textcolor{gray}{0.091} & 0.030 & \textcolor{gray}{0.124}
 & 0.469 & 0.515 & 0.473 \\
 {TTA} 
 & -0.040 & 0.125 & \textcolor{gray}{0.037}
 & 0.514 & 0.562 & 0.427 \\
 {EU} 
 & \textcolor{gray}{0.077} & 0.015 & -0.005
 & 0.473 & 0.507 & 0.427 \\
 \midrule
 \multicolumn{7}{c}{{ISIC}} \\
 \midrule
 {PriUS} 
 & \textbf{-0.478} & \textbf{0.832} & -0.492
 & \textbf{0.690} & \textbf{0.916} & 0.701 \\
 {DEviS} 
 & -0.015 & 0.044 & \textbf{-0.567}
 & 0.505 & 0.522 & \textbf{0.711} \\
 {PU} 
 & -0.032 & 0.491 & -0.502
 & 0.511 & 0.607 & 0.619 \\
 {UDrop} 
 & -0.026 & 0.112 & \textcolor{gray}{0.084}
 & 0.437 & 0.431 & 0.435 \\
 {TTA} 
 & -0.090 & 0.131 & -0.024
 & 0.488 & 0.491 & 0.478 \\
 {EU} 
 & \textcolor{gray}{0.033} & 0.077 & -0.383
 & 0.488 & 0.538 & 0.622 \\
 \midrule
 \multicolumn{7}{c}{{WHS}} \\
 \midrule
 {PriUS} 
 & \textbf{-0.501} & \textbf{0.966} & {-0.414}
 & \textbf{0.685} & \textbf{0.974} & {0.484} \\
 {DEviS} 
 & \textcolor{gray}{0.184} & 0.394 & \textbf{-0.655}
 & 0.411 & 0.679 & 0.587 \\
 {PU} 
 & \textcolor{gray}{0.055} & 0.419 & -0.494
 & 0.519 & 0.709 & 0.548 \\
 {UDrop} 
 & \textcolor{gray}{0.286} & 0.459 & -0.544
 & 0.386 & 0.729 & \textbf{0.678} \\
 {TTA} 
 & \textcolor{gray}{0.147} & 0.039 & -0.541
 & 0.441 & 0.519 & 0.633 \\
 {EU} 
 & -0.051 & 0.091 & -0.472
 & 0.517 & 0.545 & 0.569 \\

 \bottomrule
 \end{tabular}
 \label{tab:ucc_ur_comparison}
\end{table}

In general, PriUS was the only method that preserved the expected UCC sign across all three ambiguity dimensions and all evaluated datasets. 
Specifically, it showed consistent alignment with boundary contrast (\(g\)), corruption severity (\(\sigma\)), and distance to the nearest semantic boundary (\(d\)).
This advantage was most pronounced on ACDC, where PriUS achieved the strongest UCC magnitude and the highest UR on all three dimensions, while no competing method preserved all three expected signs simultaneously. The gap narrowed on ISIC, where several baselines already recovered the correct signs for contrast- and geometry-related UCC, suggesting that these ambiguity-inducing image properties were comparatively easier to capture in dermoscopic images. 
On WHS, a more demanding multi-class task, some baselines achieved stronger individual geometry-related scores, but these gains were not sustained across all three principles.
PriUS therefore remained the most balanced method overall, supporting the robustness of the proposed supervision strategy.

We next examined how this advantage manifested under each principle.

\begin{figure*}
 \centering
 \includegraphics[width=0.95\linewidth]{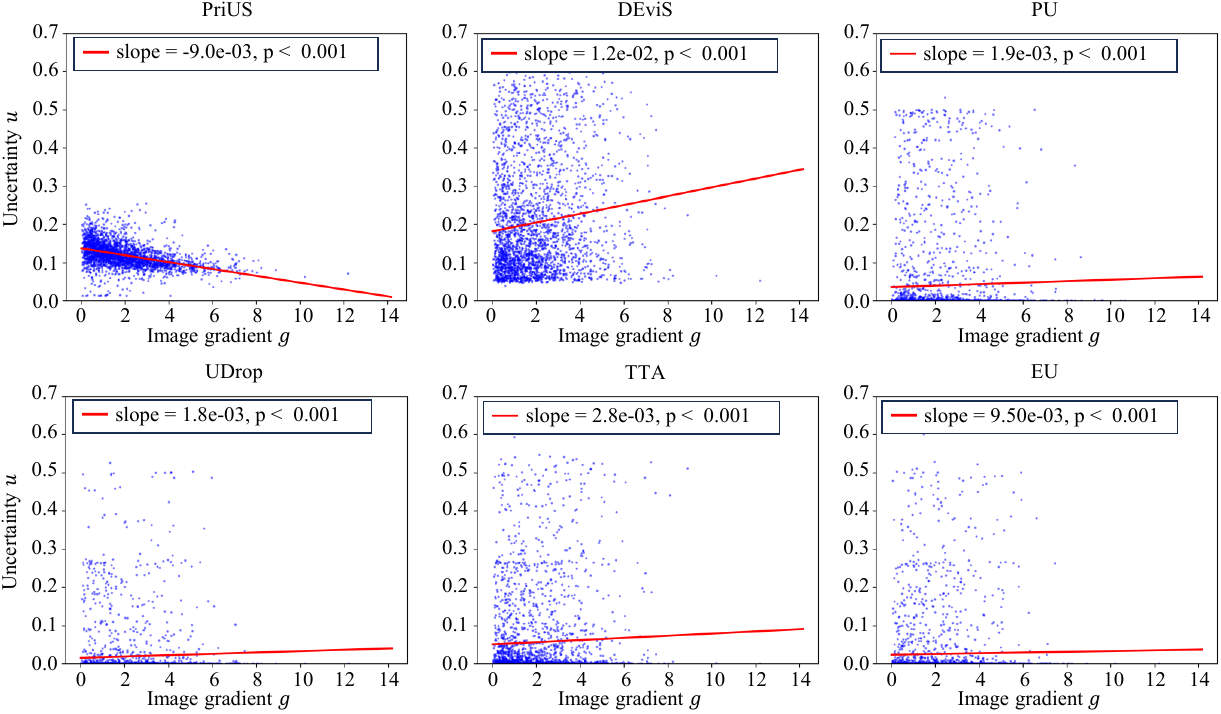}
 \caption{Contrast–uncertainty relationship at anatomical boundaries on the ACDC dataset. Scatter plots show the correlation between image gradient magnitude (a surrogate for boundary contrast) and predictive uncertainty over boundary-region pixels for each method. Each point corresponds to one evaluated boundary pixel, and the fitted regression line summarizes the overall trend. According to the Contrast Principle, a negative slope is expected, since sharper boundaries (higher image gradient magnitude) should be associated with lower uncertainty. Among all compared methods, only the proposed approach exhibited a clear negative trend, whereas competing methods showed inconsistent positive relationships.}
 \label{fig:g_u_scatter}
\end{figure*}

\textbf{Contrast.} Fig.~\ref{fig:g_u_scatter} directly supports the Contrast Principle by showing the relationship between predictive uncertainty and boundary contrast, for which the image gradient serves as a surrogate. Only PriUS exhibited a clearly negative trend between image gradient magnitude and uncertainty at boundary pixels, with a fitted regression slope of \(-9.0\times10^{-3}\). In contrast, all competing methods showed positive trends, indicating that their uncertainty estimates were inconsistent with the expected contrast-uncertainty relationship.

\begin{figure*}
 \centering
 \includegraphics[width=0.95\linewidth]{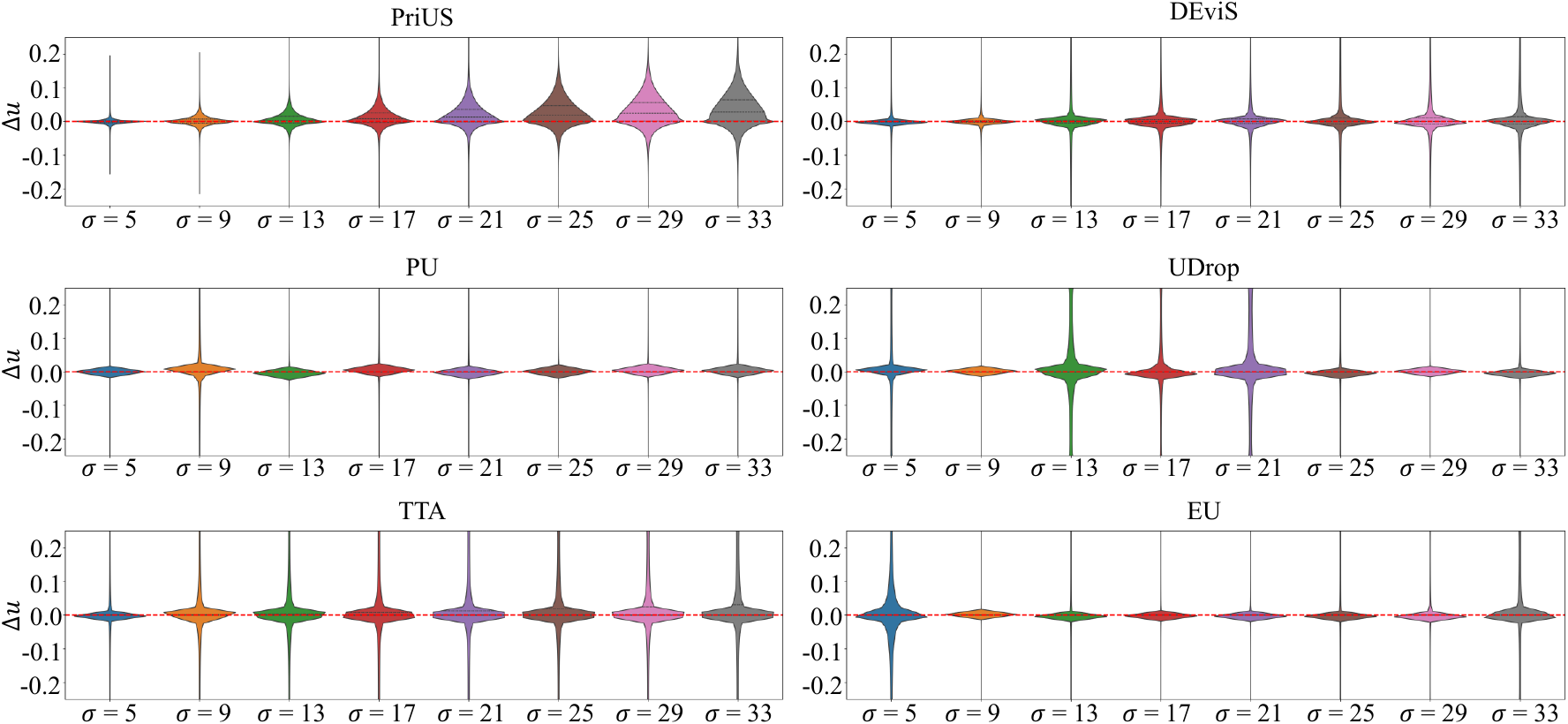}
 \caption{Uncertainty response to progressive boundary-contrast degradation induced by Gaussian blur on the ACDC dataset. 
 The change in uncertainty is defined as \(\Delta u = u^{[\sigma]} - u^{[\mathrm{original}]}\), where \(u^{[\sigma]}\) denotes the uncertainty predicted from the image with Gaussian blur level \(\sigma\), and \(u^{[\mathrm{original}]}\) denotes the uncertainty predicted from the corresponding original image. 
 Violin plots show the distributions of \(\Delta u\) at boundary-region pixels. 
 The Contrast Principle implies that decreasing boundary contrast should raise uncertainty, corresponding to \(\Delta u > 0\), i.e., values above the red dashed line \((\Delta u = 0)\). 
 PriUS yielded a substantially larger proportion of positive \(\Delta u\) values, whereas competing methods produced distributions centered near zero or approximately symmetric around zero, indicating weaker sensitivity to contrast degradation.}
 \label{fig:blur_u_change}
\end{figure*}

Fig.~\ref{fig:blur_u_change} provides additional quantitative evidence by visualizing the distributions of uncertainty change, denoted by \(\Delta u\), under increasing Gaussian blur.
As the blur level increases, boundary contrast systematically decreases. According to the Contrast Principle, uncertainty should therefore increase, making \(\Delta u > 0\) the expected behavior. Consistent with this expectation, PriUS exhibited the smallest proportion of $\Delta u < 0$ across all blur levels, indicating a stable and monotonic increase in uncertainty as boundary contrast degraded. Competing methods, by comparison, showed weak or bidirectional $\Delta u$ responses across blur kernels, suggesting that their uncertainty estimates failed to respond consistently to contrast degradation.

\begin{figure*}
 \centering
 \includegraphics[width=0.95\linewidth]{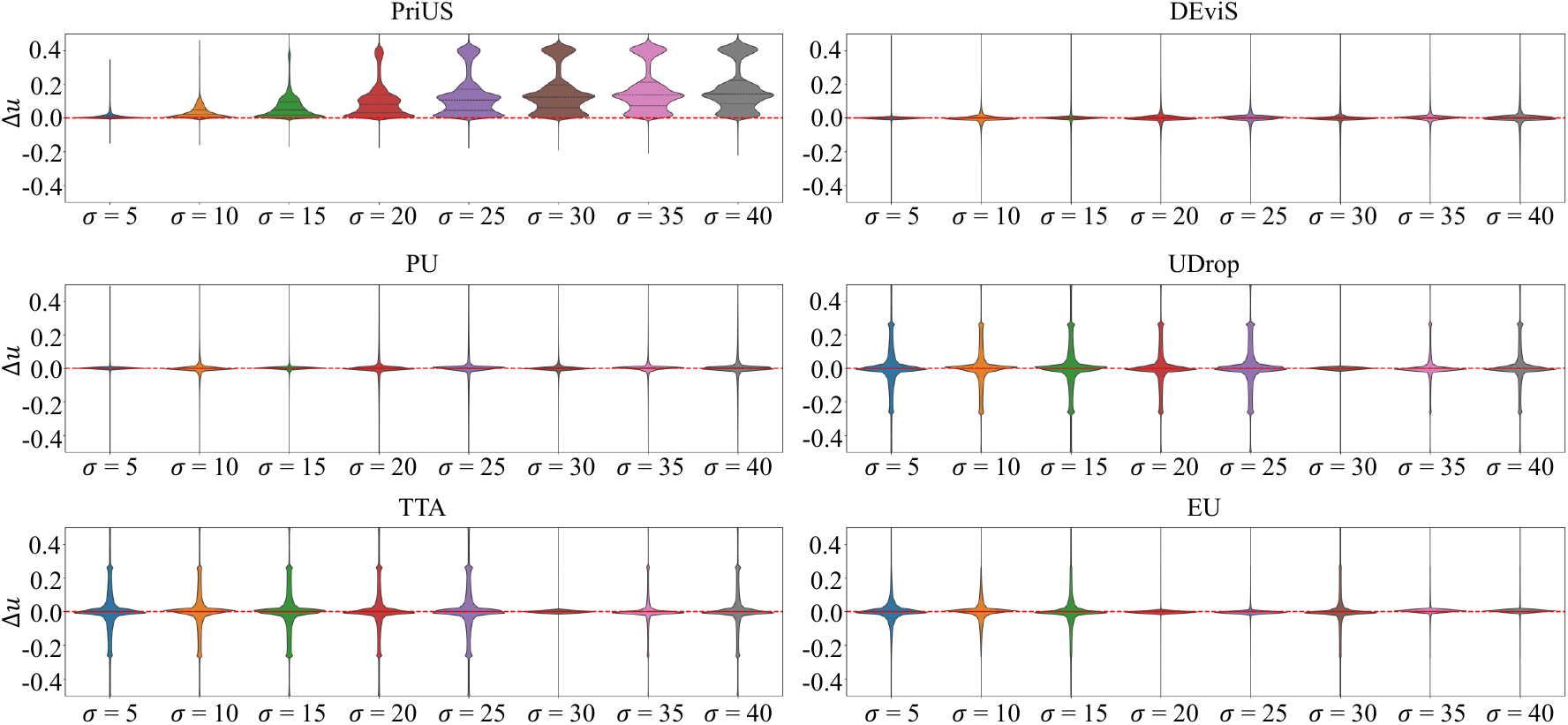}
 \caption{Uncertainty response to increasing corruption severity induced by Gaussian noise on the ACDC dataset. 
 The change in uncertainty is defined as \(\Delta u = u^{[\sigma]} - u^{[\mathrm{original}]}\), where \(u^{[\sigma]}\) denotes the uncertainty predicted from the image corrupted with Gaussian noise of level \(\sigma\), and \(u^{[\mathrm{original}]}\) denotes the uncertainty predicted from the corresponding original image without added noise. 
 Violin plots show the distributions of \(\Delta u\) at boundary-region pixels. 
 The Corruption Principle implies that increasing corruption severity should raise uncertainty, corresponding to \(\Delta u > 0\), i.e., values above the red dashed line \((\Delta u = 0)\). 
 PriUS yielded a substantially larger proportion of positive \(\Delta u\) values, whereas competing methods produced distributions centered near zero or approximately symmetric around zero, indicating weaker sensitivity to corruption severity.}
 \label{fig:noisy_u_change}
\end{figure*}

\textbf{Corruption.} Fig.~\ref{fig:noisy_u_change} examines uncertainty changes $\Delta u$ at boundary regions under increasing Gaussian noise, in line with the Corruption Principle.
As corruption severity (parameterized by $\sigma$) grew, PriUS consistently exhibited a dominant
positive shift in $\Delta u$, with the smallest proportion of $\Delta u < 0$
across all noise scales, while competing methods showed broadly symmetric or corruption-insensitive
$\Delta u$ distributions. 

\begin{figure*}
 \centering
 \includegraphics[width=0.95\linewidth]{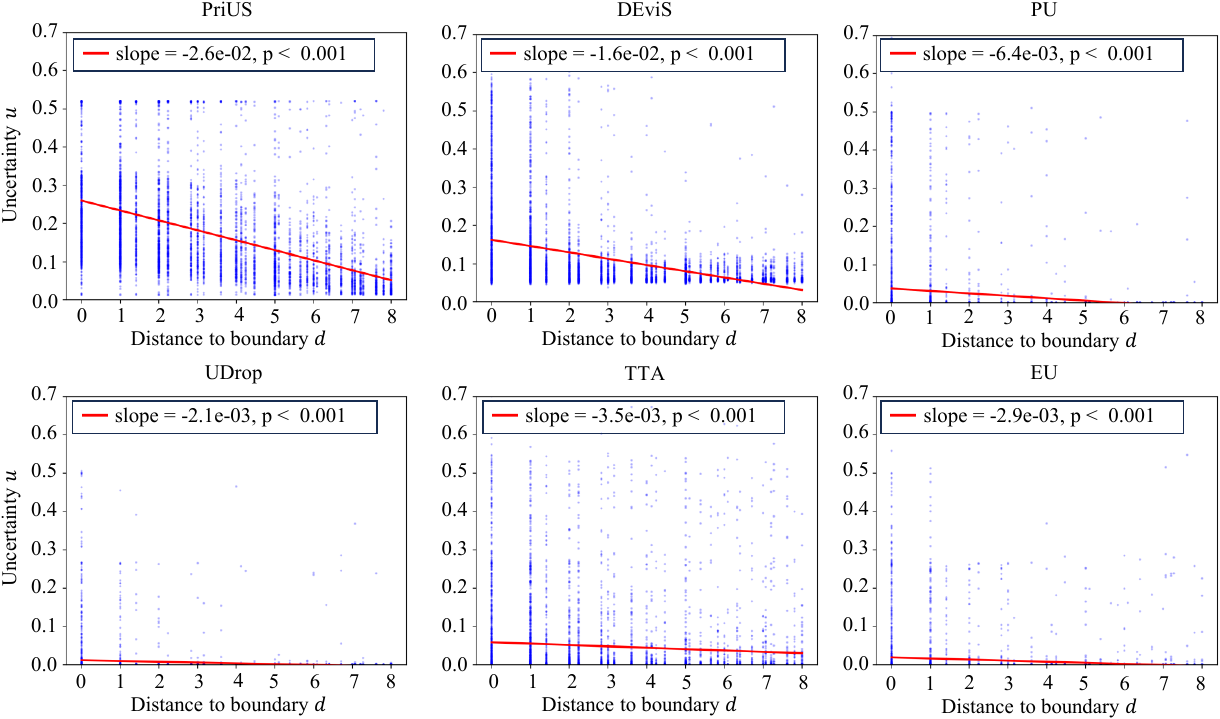} 
 \caption{Boundary distance-uncertainty relationship on the ACDC dataset. Scatter plots show the relationship between pixel-wise uncertainty and distance to the nearest class boundary, with boundary distance serving as a surrogate for geometric stability.
 Each point represents one evaluated pixel, and the fitted regression line summarizes the overall trend. According to the Geometry Principle, uncertainty is expected to decrease as the distance from ambiguous boundary regions increases. PriUS demonstrated a clear monotonic decay pattern, whereas competing methods exhibited comparatively flatter trends.}
 \label{fig:d_u_scatter}
\end{figure*}

\textbf{Geometry.}  Fig.~\ref{fig:d_u_scatter} provides quantitative support for the expected distance-uncertainty relationship by showing the correlation between uncertainty and boundary distance. PriUS exhibited a clear negative trend, with a fitted regression slope of approximately \(-2.6\times10^{-2}\), indicating that uncertainty decreased as pixels moved away from structurally ambiguous interfaces. In contrast, competing methods showed substantially weaker trends, with slopes much closer to zero, suggesting that their uncertainty estimates did not exhibit a clear or consistent decay with increasing boundary distance.

\paragraph{Qualitative analysis}
We qualitatively examined uncertainty interpretability through representative case studies organized around the three proposed principles. Specifically, we assessed whether predicted uncertainty distinguished boundaries with different local contrast, varied consistently across images with different noise levels, and exhibited the expected spatial decay from geometrically ambiguous boundary regions toward coherent interiors. Intra-image comparisons primarily revealed contrast-related behavior, inter-image comparisons primarily revealed corruption-related behavior, and the two together illustrated whether uncertainty maps remained consistent with the expected geometry-related pattern.

\begin{figure*}[th!]
 \centering
 \includegraphics[width=0.95\linewidth]{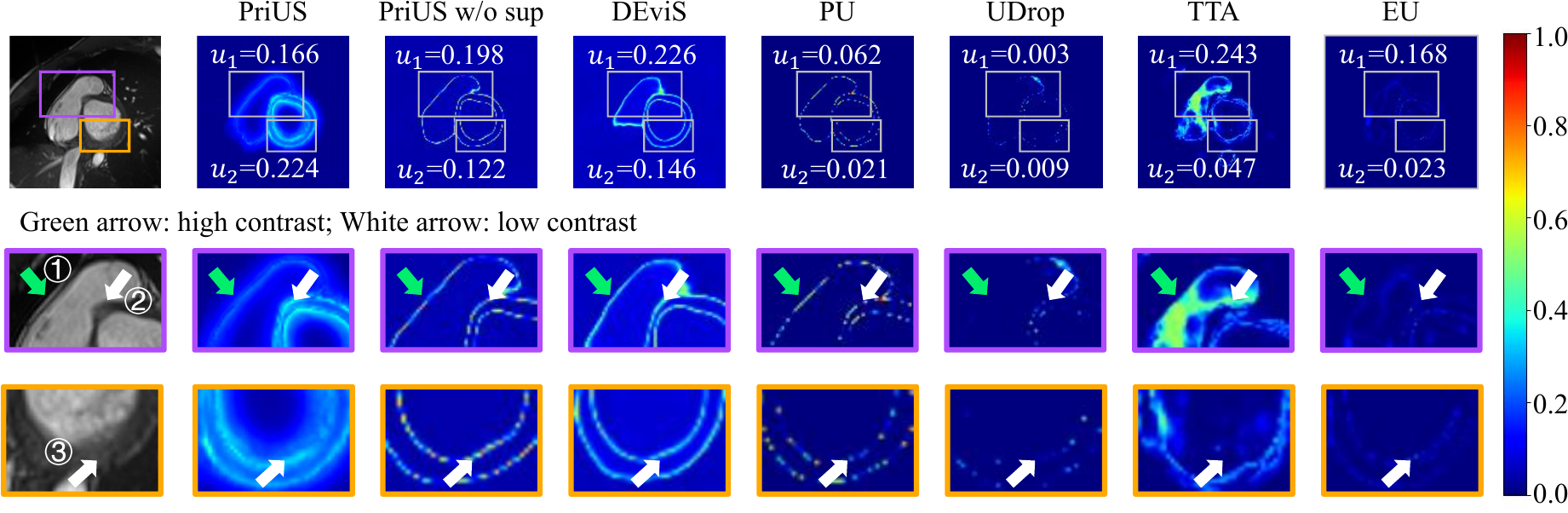}
 \caption{Intra-image comparison of uncertainty responses to boundary contrast. In the \textbf{first row}, we highlight a visually high-contrast region (purple box) and a low-contrast region (orange box) within the displayed image, and provide the averaged uncertainty values ($u_1$ and $u_2$) computed over the boundary pixels of the two regions across methods for reference. PriUS yielded noticeably lower uncertainty in the high-contrast region than in the low-contrast region, in clear agreement with the Contrast Principle. By comparison, competing methods violated this principle. In the \textbf{second row}, we zoom into the region that is overall higher in contrast. Even within this relatively high-contrast region, PriUS assigned low uncertainty to the sharp and well-defined RV blood-pool contour (arrow \ding{172}) while producing elevated uncertainty along the comparatively less distinct Myo-RV boundary (arrow \ding{173}). Competing methods failed to capture this within-region contrast-dependent variation and instead produced irregular uncertainty responses. In the \textbf{third row}, we zoom into the region that is overall lower in contrast. In this region, PriUS produced increased uncertainty along the blurred Myo-LV contour (arrow \ding{174}), whereas competing methods showed limited sensitivity to the reduced local boundary contrast.}
 \label{fig:casestudy_within}
\end{figure*}

\textbf{Contrast.} Fig.~\ref{fig:casestudy_within} shows an intra-image example of contrast-induced ambiguity. PriUS exhibited clearly differentiated uncertainty responses across boundaries with different visual clarity, indicating good alignment between the predicted uncertainty and local contrast variation. Competing methods, by comparison, provided much weaker differentiation and failed to reflect the contrast difference as clearly.

\begin{figure*}[th!]
 \centering
 \includegraphics[width=0.95\linewidth]{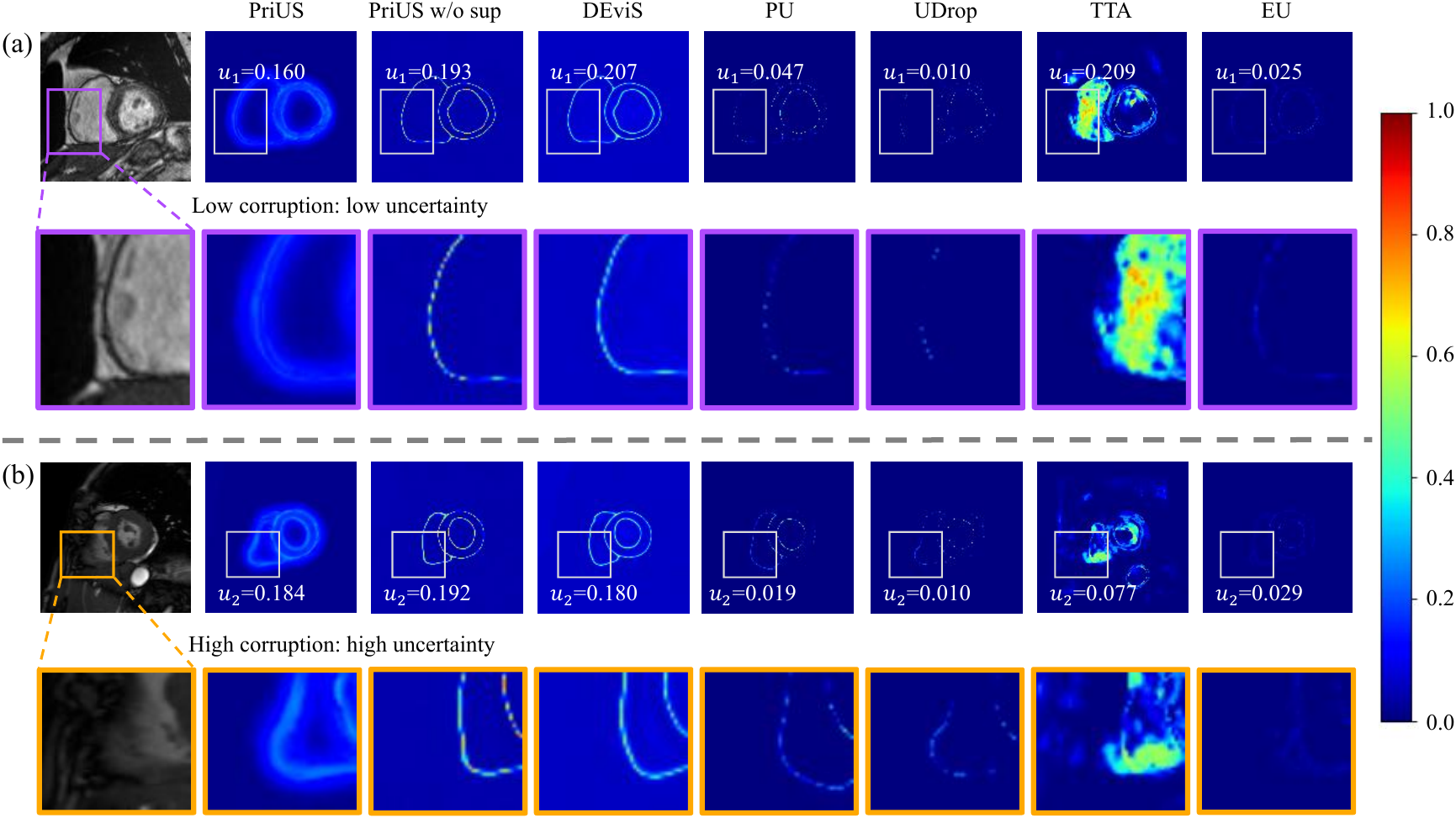}
 \caption{Inter-image uncertainty responses to global image corruption. \textbf{(a)} For a relatively less corrupted sub-image with a clear RV boundary, PriUS assigned low uncertainty. In contrast, competing methods either produced similarly high uncertainty despite the cleaner image or exhibited scattered response patterns that were not aligned with the RV interface. \textbf{(b)} For a more corrupted sub-image with less distinct RV boundaries, PriUS yielded consistently elevated uncertainty. Competing methods failed to show a coherent increase in uncertainty and instead often produced irregular or noisy responses. The averaged uncertainty values ($u_1$ and $u_2$) for the two sub-images across methods are provided for reference.}
 \label{fig:casestudy_inter}
\end{figure*}

\textbf{Corruption.} Fig.~\ref{fig:casestudy_inter} shows that PriUS responded consistently to corruption-induced ambiguity across images. The visible shift in its uncertainty pattern between less corrupted and more corrupted cases indicated that the estimated uncertainty remained sensitive to changes in image quality. Competing methods, by comparison, showed much less differentiation between the two cases.

\textbf{Geometry.} Both Fig.~\ref{fig:casestudy_within} and Fig.~\ref{fig:casestudy_inter} show geometry-related uncertainty behavior. PriUS exhibited a more gradual, spatially coherent uncertainty transition from boundary regions toward interior regions, consistent with the expected decay of structural ambiguity with increasing distance from complex interfaces. Competing methods, by comparison, concentrated uncertainty more strongly on boundary contours and provided less evidence of such a geometry-aligned transition.

\subsection{Segmentation Accuracy}
\label{subsec_segmentation}
\begin{table}[!t]
\centering
\caption{Segmentation performance comparison across ACDC, ISIC, and WHS. PriUS achieved competitive segmentation accuracy, while obtaining the best HD95 on ACDC and WHS.}
{%
\begin{tabular}{ccccccc}
\toprule
\multirow{2}{*}{Methods} 
 & \multicolumn{2}{c}{ACDC} 
 & \multicolumn{2}{c}{ISIC} 
 & \multicolumn{2}{c}{WHS} \\
\cmidrule(lr{0pt}){2-3} \cmidrule(lr{0pt}){4-5} \cmidrule(lr{0pt}){6-7}
 & \makecell{DSC \\ (\%)} & \makecell{HD95 \\ (mm)}
 & \makecell{DSC \\ (\%)} & \makecell{HD95 \\ (pixels)}
 & \makecell{DSC \\ (\%)} & \makecell{HD95 \\ (mm)} \\
\midrule
PriUS & 91.05 & \textbf{7.45} & 84.33 & 34.31 & 88.72 & \textbf{5.49} \\
DEviS & 88.74 & 9.12 & 85.68 & 25.85 & \textbf{89.80} & 5.51 \\
PU & 86.78 & 9.89 & 84.49 & 33.56 & 81.77 & 8.86 \\
UDrop & \textbf{91.36} & 8.10 & \textbf{86.20} & 24.16 & 89.14 & 6.06 \\
TTA & 70.14 & 29.02 & 83.30 & 55.66 & 87.89 & 7.18 \\
EU & 86.79 & 10.25 & 85.40 & \textbf{22.35} & 81.92 & 10.04 \\
\bottomrule
\end{tabular}}
\label{tab:dsc_hd95_comparison}
\end{table}

To verify that improved uncertainty interpretability did not come at the cost of predictive accuracy, we further evaluated segmentation performance.

Table~\ref{tab:dsc_hd95_comparison} shows that PriUS achieved competitive segmentation performance across all three datasets. In particular, it obtained the best HD95 on ACDC (7.45 mm) and WHS (5.49 mm), while maintaining DSC values comparable to the strongest competing methods. These results indicated that improving uncertainty interpretability did not compromise segmentation quality.

We attribute the HD95 advantage on ACDC and WHS to the boundary-sensitive nature of the proposed uncertainty supervision. Specifically, the contrast- and geometry-guided constraints explicitly regulate uncertainty in boundary-relevant and geometrically complex regions, where segmentation errors are more likely to cause large surface deviations. This encourages the model to better capture structurally ambiguous interfaces and thereby reduces severe boundary outliers. As HD95 is more sensitive than DSC to localized boundary errors, this benefit is reflected more clearly in HD95 than in overlap-based metrics.

\begin{figure}
 \centering
 \includegraphics[width=0.95\linewidth]{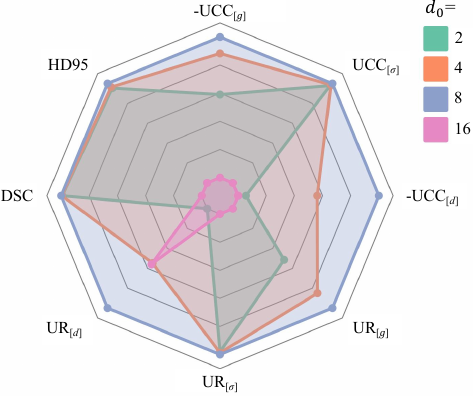}
 \caption{Parameter study of the boundary threshold $d_0$ on the ACDC dataset. All metrics were aligned to a common optimal direction (higher is better) and normalized across threshold settings. PriUS achieved the best overall trade-off at $d_0=8$, while $d_0=16$ led to training failure.}
 \label{fig:parameter_study}
\end{figure}

\begin{table*}[th!]
\centering
\caption{Ablation study on the ACDC dataset. In the UCC columns, \checkmark\ and $\times$ indicate whether the corresponding result satisfies the expected correlation sign. The \(\Delta\) entries denote the relative change with respect to the full model (PriUS). When the expected sign is violated, the corresponding \(\Delta\) value is marked as N/A, because a relative change is no longer meaningful once the uncertainty-ambiguity relationship has the wrong direction. Red and blue \(\Delta\) cells denote performance degradation or improvement, respectively.}
\setlength{\tabcolsep}{3.5pt}
\begin{tabular}{ccc|cc|cccccc|cccccc}
\toprule

\multirow{2}{*}{\textbf{$\mathcal{L}_{g}$}} 
& \multirow{2}{*}{\textbf{$\mathcal{L}_{\sigma}$}}
& \multirow{2}{*}{\textbf{$\mathcal{L}_{d}$}} 
& \multicolumn{1}{c}{DSC$\uparrow$} 
& \multicolumn{1}{c|}{HD95$\downarrow$} 
& \multicolumn{6}{c|}{UCC}
& \multicolumn{6}{c}{UR$\uparrow$} \\

\cmidrule(lr{0pt}){6-11}
\cmidrule(lr{0pt}){12-17}

& & 
& (\%) 
& (mm) 
& $g(-)$ & $\Delta_{[g]}$ & $\sigma(+)$ & $\Delta_{[\sigma]}$ & $d(-)$ & $\Delta_{[d]}$
& $g$ & $\Delta_{[g]}$ & $\sigma$ & $\Delta_{[\sigma]}$ & $d$ & $\Delta_{[d]}$ \\ 

\midrule

\checkmark & \checkmark & \checkmark
& {91.05} & \textbf{7.45}
& {-0.597} & \basedelta
& {0.973} & \basedelta
& {-0.767} & \basedelta
& {0.708} & \basedelta
& {0.986} & \basedelta
& {0.739} & \basedelta \\

\cdashline{1-17}

{$\times$} & \checkmark & \checkmark
& {89.39} & {11.15}
& \textcolor{red}{$\times$} & {N/A}
& {\checkmark} & \negdelta{10}{0.006}
& {\checkmark} & \posdelta{11}{0.031}
& {0.446} & \negdelta{18}{0.262}
& {0.983} & \negdelta{8}{0.003}
& {0.759} & \posdelta{8}{0.020} \\

\checkmark & {$\times$} & \checkmark
& {90.65} & {10.66}
& {\checkmark} & \negdelta{12}{0.090}
& \textcolor{red}{$\times$} & {N/A}
& {\checkmark} & \negdelta{15}{0.128}
& {0.674} & \negdelta{9}{0.034}
& {0.450} & \negdelta{27}{0.536}
& {0.700} & \negdelta{10}{0.039} \\

\checkmark & \checkmark & {$\times$}
& {54.67} & {31.79}
& {\checkmark} & \posdelta{14}{0.170}
& {\checkmark} & \negdelta{20}{0.616}
& {\checkmark} & \negdelta{28}{0.501}
& {0.788} & \posdelta{10}{0.080}
& {0.636} & \negdelta{20}{0.350}
& {0.524} & \negdelta{16}{0.215} \\

\checkmark & {$\times$} & {$\times$}
& \textbf{91.21} & {8.55}
& {\checkmark} & \posdelta{16}{0.242}
& {\checkmark} & \negdelta{25}{0.909}
& {\checkmark} & \negdelta{14}{0.111}
& {0.836} & \posdelta{13}{0.128}
& {0.532} & \negdelta{23}{0.454}
& {0.652} & \negdelta{11}{0.087} \\

{$\times$} & \checkmark & {$\times$}
& {72.78} & {22.57}
& {\checkmark} & \negdelta{24}{0.590}
& {\checkmark} & \negdelta{19}{0.515}
& {\checkmark} & \negdelta{19}{0.240}
& {0.486} & \negdelta{16}{0.222}
& {0.729} & \negdelta{17}{0.257}
& {0.623} & \negdelta{12}{0.116} \\

{$\times$} & {$\times$} & \checkmark
& {90.65} & {9.75}
& \textcolor{red}{$\times$} & {N/A}
& {\checkmark} & \negdelta{24}{0.841}
& {\checkmark} & \posdelta{11}{0.035}
& {0.435} & \negdelta{19}{0.273}
& {0.566} & \negdelta{22}{0.420}
& {0.759} & \posdelta{8}{0.020} \\

{$\times$} & {$\times$} & {$\times$}
& {91.07} & {9.03}
& \textcolor{red}{$\times$} & {N/A}
& \textcolor{red}{$\times$} & {N/A}
& {\checkmark} & \negdelta{23}{0.361}
& {0.448} & \negdelta{18}{0.260}
& {0.471} & \negdelta{26}{0.515}
& {0.531} & \negdelta{16}{0.208} \\

\bottomrule
\end{tabular}
\label{tab:model_ablation}
\end{table*}

\subsection{Parameter Study}
\label{subsec_parameter}

To assess the effect of the boundary-based gating function, we investigated how the boundary threshold $d_0$ influenced overall performance on ACDC. This threshold was defined as $d_0 = d_n + \delta = d_f - \delta$, such that varying $d_0$ simultaneously adjusted the corruption-related threshold $d_n$ and the geometry-based supervision threshold $d_f$. We considered four values, $d_0 \in \{2, 4, 8, 16\}$, to examine the sensitivity of the method to this threshold and determine the most effective setting.

For joint visualization, all metrics were first aligned to a common optimal direction so that higher values consistently indicated better performance. Metrics such as HD95, for which lower values are better, were reversed accordingly and then normalized across threshold settings. This normalization ensured a consistent scale while preserving relative performance trends.

Fig.~\ref{fig:parameter_study} shows that increasing $d_0$ from 2 to 8 led to consistent improvements across most metrics, indicating that incorporating a broader boundary region provided more informative supervision.
The most pronounced gains were observed in the geometry-related uncertainty metrics, particularly $\text{UCC}_{[d]}$ and $\text{UR}_{[d]}$, which improved monotonically with $d_0$. This suggested that a larger boundary neighborhood more effectively enforced geometry-related uncertainty ordering. Moderate but consistent improvements were also observed for the contrast-related measures $\text{UCC}_{[g]}$ and $\text{UR}_{[g]}$, while other metrics stayed relatively stable within this range.

Further increasing the threshold to $d_0=16$ led to training failure, characterized by severe degradation in segmentation performance (DSC = 25.75\% and HD95 = 54.15 mm). This indicates that an excessively large boundary region dilutes semantic supervision and destabilizes optimization.

Overall, $d_0 = 8$ offered the best trade-off between segmentation accuracy and uncertainty quality on ACDC and was therefore adopted in our experiments. We note that the optimal boundary threshold may vary across datasets due to differences in image resolution, anatomy, and boundary characteristics.

\subsection{Ablation Study}
\label{subsec_ablation}

To examine the contribution of each supervision component, we conducted an ablation study on the ACDC dataset.

Table~\ref{tab:model_ablation} shows that the full model, which jointly incorporated $\mathcal{L}_{g}$, $\mathcal{L}_{\sigma}$, and $\mathcal{L}_{d}$, was the only configuration that simultaneously achieved strong segmentation performance and preserved the expected uncertainty behavior across all three principles. It attained a DSC of 91.05\%, the best HD95 of 7.45 mm, the correct UCC signs, and the strongest overall UR performance.

Removing $\mathcal{L}_{g}$ caused the contrast-related uncertainty behavior to break down: $\text{UCC}_{[g]}$ flipped from the expected negative correlation to a positive one, and $\text{UR}_{[g]}$ also dropped substantially. Similarly, removing $\mathcal{L}_{\sigma}$ reversed the expected corruption-related trend, with $\text{UCC}_{[\sigma]}$ becoming negative and $\text{UR}_{[\sigma]}$ decreasing markedly. These results indicate that $\mathcal{L}_{g}$ and $\mathcal{L}_{\sigma}$ are respectively necessary for aligning uncertainty with boundary contrast and corruption severity.

By contrast, removing $\mathcal{L}_{d}$ led to the most severe deterioration in segmentation quality, with DSC dropping to 54.67\% and HD95 increasing to 31.79 mm. This suggested that geometry-based supervision was critical not only for geometry-consistent uncertainty behavior but also for preserving stable structural prediction. Although the geometry-related UCC sign remained correct, both its magnitude and the corresponding UR value degraded, indicating weaker geometry-aligned uncertainty behavior.

The remaining partial combinations further confirmed that no individual term or incomplete combination could replace the full design. Some variants retained competitive DSC or preserved the expected sign for a subset of metrics, but they either failed to maintain correct uncertainty behavior across all principles or suffered from inferior segmentation quality.

Overall, the ablation results showed that the three supervision terms were complementary. Specifically, $\mathcal{L}_{g}$ primarily enforced contrast-aligned uncertainty, $\mathcal{L}_{\sigma}$ preserved corruption-consistent uncertainty responses, and $\mathcal{L}_{d}$ was essential for stable structural prediction while also supporting geometry-aligned uncertainty. Only their joint use produced uncertainty estimates that were simultaneously interpretable, well-balanced, and compatible with strong segmentation performance.

\section{Conclusion}
\label{sec:conclusion}

In this work, we have formulated three perception-aligned principles for interpretable uncertainty in medical image segmentation, describing how predictive uncertainty should vary with boundary contrast, image corruption, and anatomical geometry. 
Guided by these principles, we developed PriUS, a principle-guided uncertainty supervision framework instantiated within evidential learning to regularize uncertainty through principle-specific image-derived proxies.
Experiments on multiple benchmarks demonstrated that PriUS achieved reliable segmentation performance while producing more interpretable uncertainty estimates.

Unlike conventional approaches that treat uncertainty largely as a by-product of prediction, PriUS aligns uncertainty responses with clinically meaningful structural and observational cues through structured supervision. Consequently, the resulting uncertainty maps exhibit improved interpretability and are better positioned to support informed decision-making in medical image analysis.

Future work will consider more expressive ranking formulations for uncertainty supervision and extend the proposed principles to additional imaging modalities and clinical tasks. 
These efforts will also include evaluating PriUS with uncertainty estimators other than evidential learning and developing data-driven strategies for deriving additional principles, thereby improving the adaptability of the framework.

\section*{Acknowledgments}
This work was funded by the Science and Technology Commission of Shanghai Municipality (25TS1412100),
the Noncommunicable Chronic Disease-National Science and Technology Major Project (2026ZD0555800/2026ZD0555802), 
and the National Natural Science Foundation of China (62372115).

\bibliographystyle{IEEEtran}
\bibliography{ref}

\begin{IEEEbiography}
[{\includegraphics[width=1in,height=1.25in,clip,keepaspectratio]{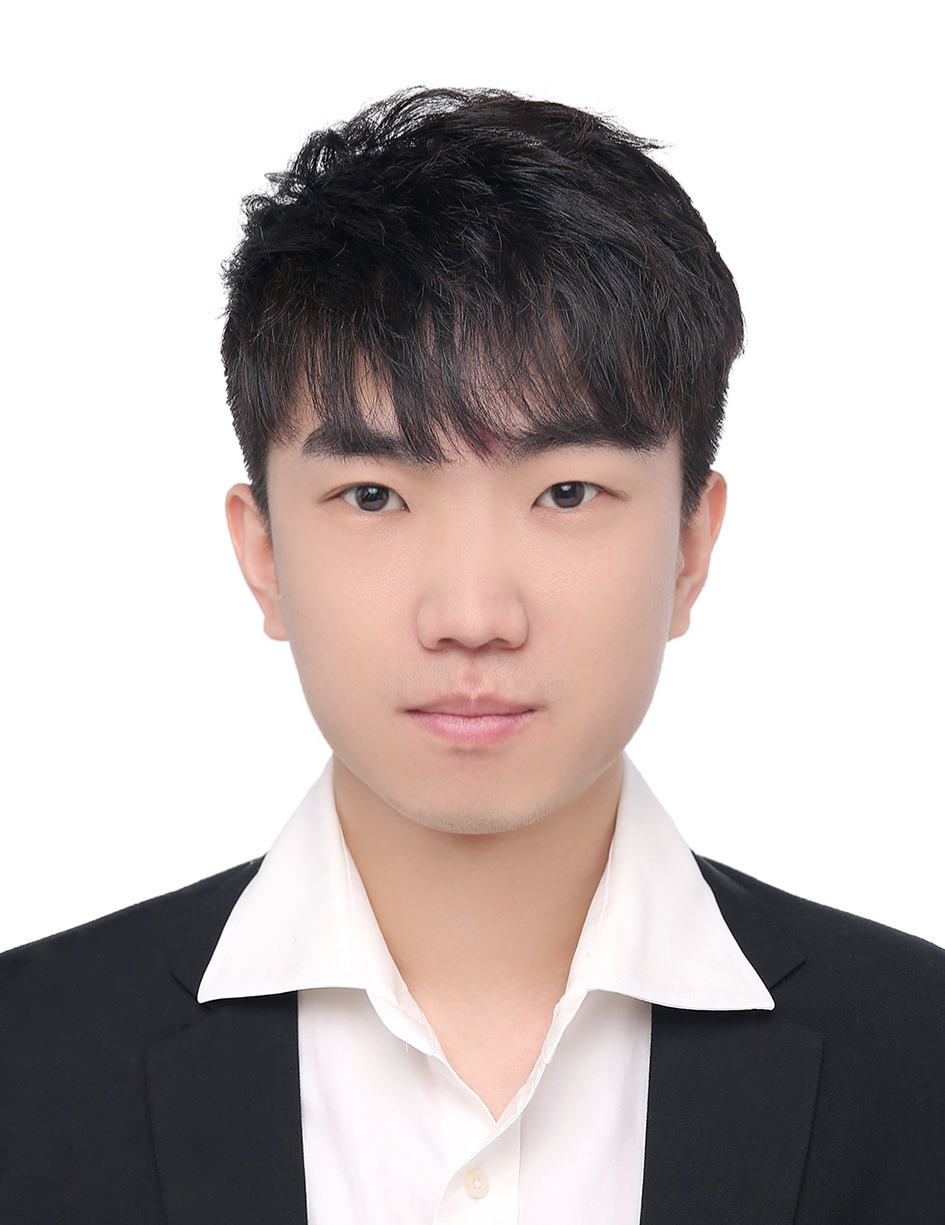}}]
{An Sui}
is currently a Ph.D. candidate in Statistics at the School of Data Science, Fudan University, Shanghai, China, supervised by Prof. Xiahai Zhuang. He received the B.S. degree in Electronic Information Science and Technology and the M.Sc. degree in Biomedical Engineering from Fudan University in 2020 and 2023, respectively. His research interests include interpretable AI, medical image analysis, and uncertainty estimation.
\end{IEEEbiography}

\begin{IEEEbiography}
[{\includegraphics[width=1in,height=1.25in,clip,keepaspectratio]{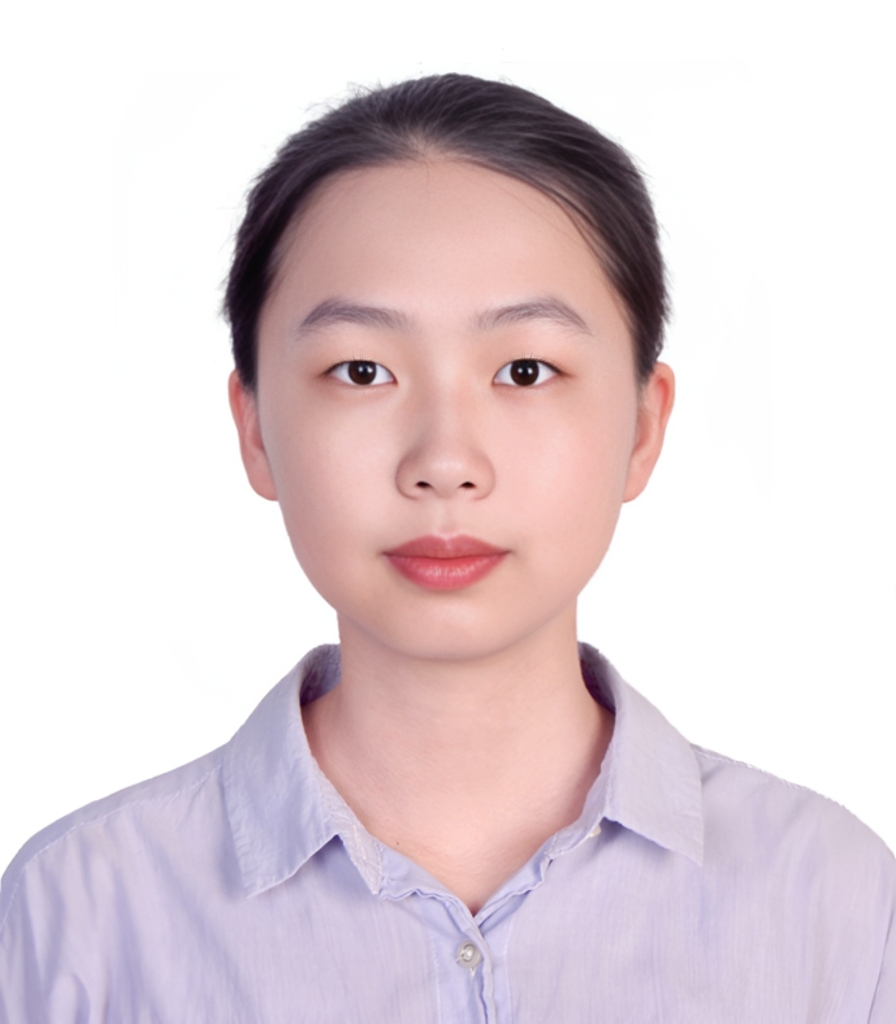}}]
{Yuzhu Li} is currently
a Ph.D. candidate in Biomedical Engineering at the Institute of Science and Technology for Brain-Inspired Intelligence (ISTBI), Fudan University, jointly supervised by Prof. Gunter Schumann and Prof. Xiahai Zhuang. She received the B.S. degree from the School of Computer Science and Engineering, Northeastern University in 2023. Her research interests include interpretable AI, medical image analysis and multi-omics analysis.
\end{IEEEbiography}

\begin{IEEEbiography}
[{\includegraphics[width=1in,height=1.25in,clip,keepaspectratio]{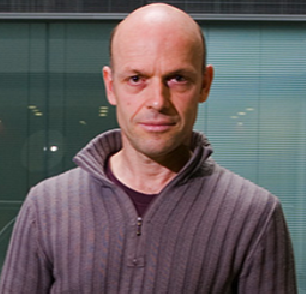
}}] {Gunter Schumann} is a professor at the ISTBI, Fudan University, and Chair and Director of the Centre for Population Neuroscience and Stratified Medicine. He is also a professor in the Department of Psychiatry and Neuroscience, Charité University Medicine. He received the Ph.D. degree from the University of Hamburg in 1994. His research interests include neuroimaging analysis and interpretation, genetics, addiction biology, and pharmacogenetics.

\end{IEEEbiography}

\begin{IEEEbiography}[{\includegraphics[width=1in,height=1.25in,clip,keepaspectratio]{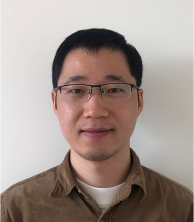}}]{Fuping Wu} is currently a postdoctoral researcher at the Imperial College London. He received the Ph.D. degree from Fudan University in 2022, the M.Sc. degree from the School of Mathematics and Statistics, Wuhan University, in 2016, and the B.S. degree from Tongji Medical College, Huazhong University of Science and Technology, in 2012. His research interests include medical image analysis, domain adaptation, and foundation models.
\end{IEEEbiography}

\begin{IEEEbiography}[{\includegraphics[width=1in,height=1.25in,clip,keepaspectratio]{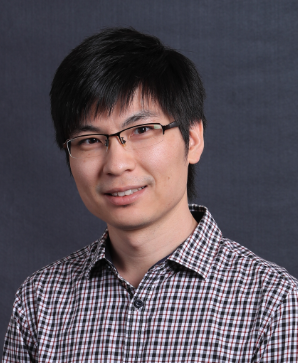}}]{Xiahai Zhuang}
is a professor at the School of Data Science, Fudan University. He received the B.S. degree from the Department of Computer Science, Tianjin University, the M.Sc. degree from Shanghai Jiao Tong University, and the Ph.D. degree from University College London. His research interests include interpretable AI, medical image analysis and computer vision. His work won the Elsevier-MedIA 1st Prize and Medical Image Analysis MICCAI Best Paper Award 2023.
\end{IEEEbiography}

\end{document}